\documentclass[11pt,a4paper]{article}
\usepackage[hyperref]{emnlp2020}
\usepackage{times}
\usepackage{latexsym}

\usepackage{microtype}

\usepackage{graphicx}
\usepackage{amsfonts}
\usepackage{amsmath}
\usepackage{amssymb}
\usepackage{pifont}
\usepackage{nicefrac}
\usepackage{multirow}
\usepackage{subcaption}
\usepackage[font=small]{caption}
\usepackage{soul}
\usepackage{xcolor}
\usepackage{booktabs}
\usepackage{algorithm}
\usepackage{algpseudocode}
\usepackage{mathtools}
\usepackage{comment}
\usepackage{textcomp}

\usepackage{tablefootnote}

\definecolor{bblue}{HTML}{4F81BD}
\definecolor{rred}{HTML}{c4260b}
\definecolor{ggreen}{HTML}{098c1f}
\definecolor{ppurple}{HTML}{9F4C7C}
\definecolor{oorange}{HTML}{F79646}

\algnewcommand\algorithmicinput{\textbf{Input:}}
\algnewcommand\Input{\item[\algorithmicinput]}
\algnewcommand\algorithmicoutput{\textbf{Output:}}
\algnewcommand\Output{\item[\algorithmicoutput]}

\algnewcommand\algorithmicempty{~}
\algnewcommand\Empty{\item[\algorithmicempty]}

\DeclarePairedDelimiter\ceil{\lceil}{\rceil}
\DeclarePairedDelimiter\floor{\lfloor}{\rfloor}

\usepackage{enumitem}
\setenumerate[1]{itemsep=0.30pt,partopsep=0pt,parsep=\parskip,topsep=5pt}
\setitemize[1]{itemsep=0.30pt,partopsep=00pt,parsep=\parskip,topsep=5pt}
\setdescription{itemsep=0.30pt,partopsep=0pt,parsep=\parskip,topsep=5pt}

\usepackage{todonotes}

\aclfinalcopy 

\newcommand\bidecoder{IBDecoder}
\newcommand\midecoder{IMDecoder}
\newcommand\sa{SA}

\title{Fast Interleaved Bidirectional Sequence Generation}

\author{Biao Zhang$^1$ \quad Ivan Titov$^{1,2}$ \quad Rico Sennrich$^{3,1}$ \bigskip\\
  $^1$School of Informatics, University of Edinburgh \\
  $^2$ILLC, University of Amsterdam \\
  $^3$Department of Computational Linguistics, University of Zurich \\
  \texttt{B.Zhang@ed.ac.uk, ititov@inf.ed.ac.uk, sennrich@cl.uzh.ch}
  }

\date{}

\begin{document}
\maketitle
\begin{abstract}

Independence assumptions during sequence generation can speed up inference, but parallel generation of highly inter-dependent tokens comes at a cost in quality.
Instead of assuming independence between neighbouring tokens (semi-autoregressive decoding, \sa{}), we take inspiration from bidirectional sequence generation and introduce a decoder that generates target words from the left-to-right and right-to-left directions simultaneously.
We show that we can easily convert a standard architecture for unidirectional decoding into a bidirectional decoder by simply interleaving the two directions and adapting the word positions and self-attention masks. Our interleaved bidirectional decoder \mbox{(\bidecoder{})} retains the model simplicity and training efficiency of the standard Transformer, and on five machine translation tasks and two document summarization tasks, achieves a decoding speedup of $\sim$2$\times$ compared to autoregressive decoding with comparable quality. Notably, it outperforms left-to-right \sa{} because the independence assumptions in \bidecoder{} are more felicitous.
To achieve even higher speedups, we explore hybrid models where we either simultaneously predict multiple neighbouring tokens per direction, or perform multi-directional decoding by partitioning the target sequence. These methods achieve speedups to 4$\times$--11$\times$ across different tasks at the cost of $<$1 BLEU or $<$0.5 ROUGE (on average).\footnote{Source code is released at \url{https://github.com/bzhangGo/zero}.}

\end{abstract}

\section{Introduction}

Neural sequence generation aided by encoder-decoder models~\cite{DBLP:journals/corr/BahdanauCB14,NIPS2017_7181} has achieved great success in recent years~\cite{bojar-etal-2018-findings,00252,2019t5,9003750}, but still suffers from slow inference. One crucial bottleneck lies in its generative paradigm which factorizes the conditional probability along the target sequence $\mathbf{y}=\{y_1, y_2, \ldots, y_n\}$ of length $n$ as follows:
\begin{equation}\label{eq:standard_prob}
    p(\mathbf{y}|\mathbf{x}) = \prod_{t=1}^{n} p\left(y_t|\mathbf{y}_{<t}, \mathbf{x}\right),
\end{equation}
where $\mathbf{x}=\{x_1, x_2, \ldots, x_m\}$ is the source sequence of length $m$. This factorization determines that target words can only be generated one-by-one in a sequential and unidirectional manner, which limits the decoding efficiency.

\begin{figure*}[t]
  \centering

    \subcaptionbox{\label{fig:overall:bidecoder}\bidecoder{}, $h=2, c=1$}{
        \includegraphics[scale=0.43]{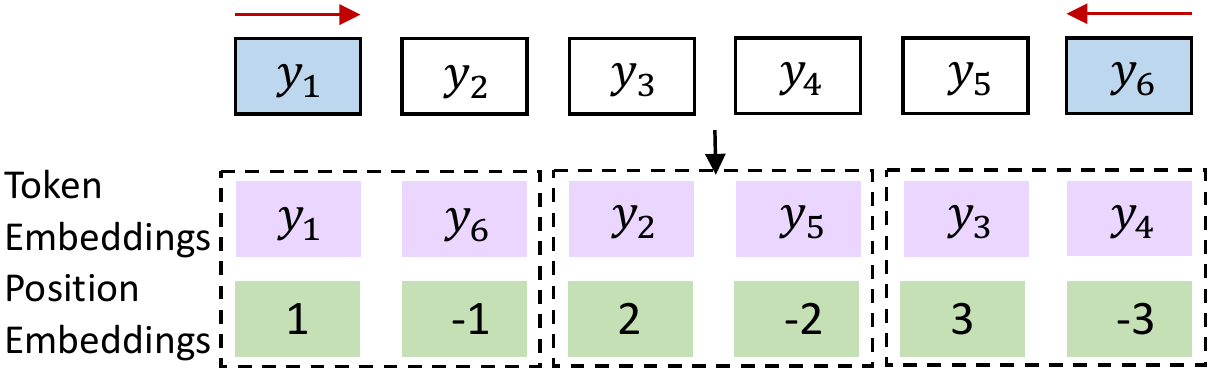}
    }%
    \subcaptionbox{\label{fig:overall:sa}\sa{}, $h=1, c=2$}{
        \includegraphics[scale=0.43]{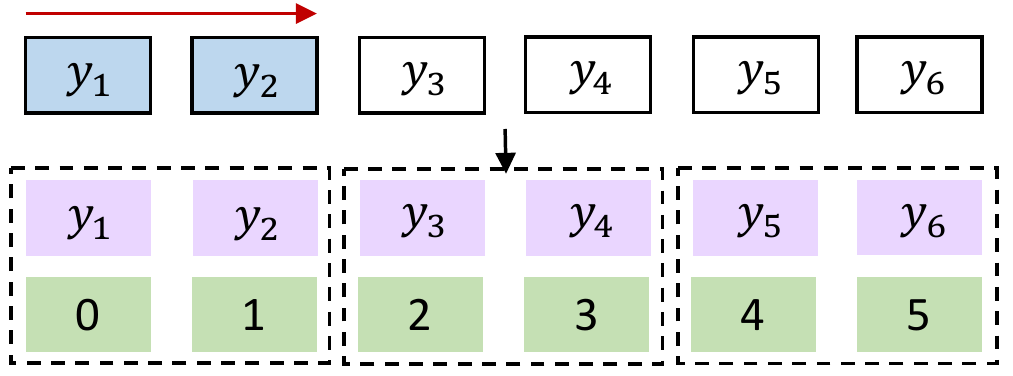}
    }~%
    \subcaptionbox{\label{fig:overall:mask2} Att. Mask for (\ref{fig:overall:bidecoder}, \ref{fig:overall:sa})}[3.5cm]{
        \includegraphics[scale=0.43]{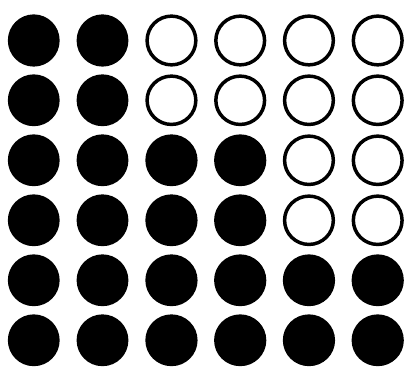}
    }
    \\
    \subcaptionbox{\label{fig:overall:midecoder}\midecoder{}, $h=4, c=1$}{
        \includegraphics[scale=0.43]{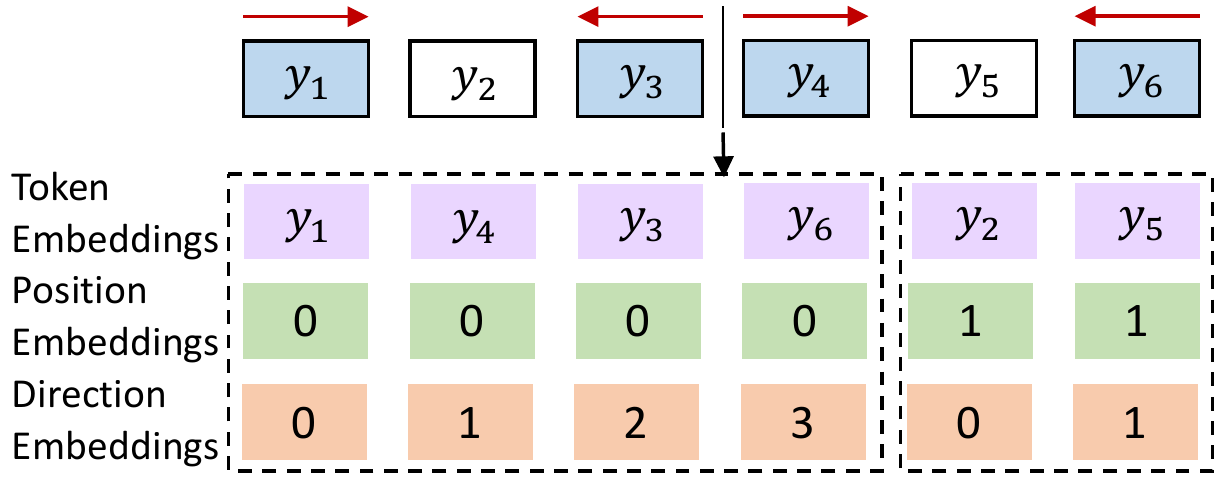}
    }~%
    \subcaptionbox{\label{fig:overall:bidecoder_sa}\bidecoder{} + \sa{}, $h=2, c=2$}{
        \includegraphics[scale=0.43]{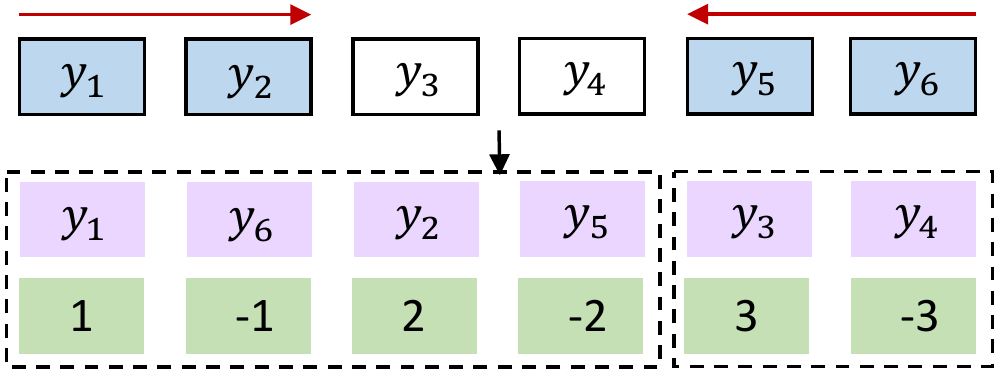}
        \vspace{0.46cm}
    }~%
    \subcaptionbox{\label{fig:overall:mask4} Att. Mask for (\ref{fig:overall:midecoder}, \ref{fig:overall:bidecoder_sa})}[3.5cm]{
        \includegraphics[scale=0.43]{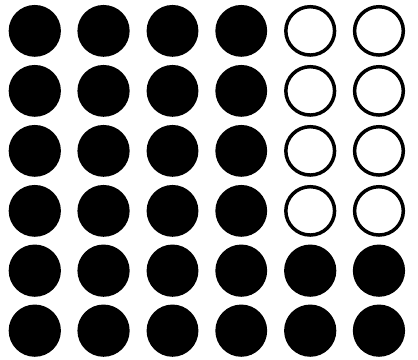}
    }%
    
  \caption{\label{fig:overall} Overview of the interleaved bidirectional decoder (\bidecoder, \ref{fig:overall:bidecoder}), the semi-autoregressive decoder (\sa, \ref{fig:overall:sa}), the interleaved multi-directional decoder (\midecoder, \ref{fig:overall:midecoder}) and the bidirectional semi-autoregressive decoder (\bidecoder + \sa, \ref{fig:overall:bidecoder_sa}) on target sequence $\mathbf{y}=\{y_1, y_2, \ldots, y_6\}$.
  We reorganize the target sequence (purple), the word positions (green) and the self-attention mask (circles) to reuse the standard Transformer decoder. During inference, multiple target words are generated simultaneously at each step (dashed rectangles), improving the decoding speed.
  The self-attention masks are given in (\ref{fig:overall:mask2}) and (\ref{fig:overall:mask4}), where sold black circles indicate allowed attention positions.
  Red arrows indicate generation directions ($h$ is the direction number), whose length denotes the number of words produced per direction ($c$). Blue rectangles denote words generated at the first step.
  The direction embedding (red rectangles) reflects the direction each target word belongs to. 
  Apart from the left-to-right generation, \bidecoder{} jointly models the right-to-left counterpart within a single sequence. \midecoder{} extends \bidecoder{} by splitting the sequence into several equal segments and performing bidirectional generation on each of them, while \bidecoder+\sa{} allows each direction to produce multiple words.
 }
\end{figure*}

A promising direction to break this barrier is to generate multiple target words at one decoding step to improve the parallelization of inference~\cite{gu2018nonautoregressive,NIPS2018_8212}.
However, this introduces independence assumptions that hurt translation quality, since words produced in parallel are in fact likely to be inter-dependent.
We hypothesize that there are groups of words that are less likely to be strongly inter-dependent than neighbouring words, which will allow for better parallelization.
Inspired by bidirectional modeling~\cite{8863411,ZHANG2020103234}, we resort to an alternative probabilistic factorization:
\begin{equation}\label{eq:two_head_prob}
    p^{\textsc{bd}}(\mathbf{y}|\mathbf{x}) = \prod_{t=1}^{\ceil*{\nicefrac{n}{2}}} p\left(\overrightarrow{y_t}, \overleftarrow{y_{t^\prime}}|\overrightarrow{\mathbf{y}_{<t}}, \overleftarrow{\mathbf{y}_{>{t^\prime}}}, \mathbf{x}\right),
\end{equation}
Introducing an independence assumption between $t$ and $t^\prime =n-t+1$ allows for parallel word prediction from both the $\overrightarrow{\text{left-to-right}}$ and $\overleftarrow{\text{right-to-left}}$ directions.
Based on this factorization, \citet{zhou2019sequence} propose synchronous bidirectional translation using a dedicated interactive decoder, and report quality improvements compared to left-to-right semi-autoregressive decoding~\cite[\sa{}]{wang-etal-2018-semi} in translation quality. However, their success comes along with extra computational overhead brought by the specialized decoder.
Empirically, \citet{zhou2019sequence} only report a decoding speedup of 1.38$\times$, slower than \sa{}, although the factorization halves the decoding steps.

We combine the strengths of bidirectional modeling and \sa{}, and propose interleaved bidirectional decoder (\bidecoder{}) for fast generation. As shown in Figure \ref{fig:overall:bidecoder}, we interleave target words from the left-to-right and right-to-left directions and separate their positions to support reusing any standard unidirectional decoders, such as the Transformer decoder~\cite{NIPS2017_7181}. We reorganize the self-attention mask to enable inter- and intra-direction interaction (Figure \ref{fig:overall:mask2}) following \sa{}. Unlike \sa{}, we show through experiments that distant tokens from different directions are less inter-dependent, providing a guarantee for better performance. Compared to previous studies~\cite{zhang2018asynchronous,8863411,ZHANG2020103234,zhou2019sequence}, our approach has no extra model parameters and brings in little overhead at training and decoding. 

\bidecoder{} is speedup-bounded at 2$\times$. To push this ceiling up, we explore strategies for multi-word simultaneous generation, including \textit{multi-directional decoding} (\midecoder, Figure \ref{fig:overall:midecoder}) and \textit{\sa{}} (Figure \ref{fig:overall:sa}). The former extends Eq. \ref{eq:two_head_prob} by inserting more generation directions, while the latter allows each direction to produce multiple target words~\cite{wang-etal-2018-semi}. These strategies offer us a chance to aggressively improve the decoding speed albeit at the risk of degenerated performance. To encourage multi-word generation in parallel, we propose a modified beam search algorithm.

We extensively experiment on five machine translation tasks and two document summarization tasks, with an in-depth analysis studying the impact of batch size, beam size and sequence length on the decoding speed. We close our analysis by examining the capacity of our model in handling long-range dependencies. 
On these tasks, \bidecoder{} yields $\sim$2$\times$ speedup against Transformer at inference, and reaches 4$\times$--11$\times$ after pairing it with \sa{}. Still, the overall generation quality is comparable. When we pair our method with sequence-level knowledge distillation \citep{kim-rush-2016-sequence}, we outperform a Transformer baseline on 6 out of 7 tasks.

Our contributions are summarized below:
\begin{itemize}
    \item We propose \bidecoder{}, following a bidirectional factorization of the conditional probability, for fast sequence generation. \bidecoder{} retains the training efficiency and is easy to implement.
    \item We extend \bidecoder{} to enable multi-word simultaneous generation by investigating integration with \midecoder{} and \sa{}. Results show that \bidecoder{} + \sa{} performs better than \midecoder{}.
    \item We propose a modified beam search algorithm to support step-wise parallel generation.
    \item On several sequence generation benchmarks, \bidecoder{} yields $\sim$2$\times$ speedup against Transformer at inference, and reaches 4$\times$--11$\times$ after pairing it with \sa{}. Still, the overall generation quality is comparable. 
\end{itemize}

\section{Related Work}

Efforts on fast sequence generation come along with the rapid development of encoder-decoder models~\cite{NIPS2017_7181}. A straightforward way is to reduce the amount of computation. Methods in this category range from teacher-student model~\cite{kim-rush-2016-sequence,hayashi-etal-2019-findings}, constrained softmax prediction~\cite{hu2015improved}, beam search cube pruning~\cite{zhang-etal-2018-speeding}, float-point quantization~\cite{wu2016google,bhandare2019efficient}, model pruning~\cite{see-etal-2016-compression}, to simplified decoder architectures, such as lightweight recurrent models~\cite{zhang-etal-2018-simplifying,zhang-sennrich-2019-lightweight,kim-etal-2019-research}, average attention network~\cite{zhang-etal-2018-accelerating}, merged attention network~\cite{zhang-etal-2019-improving}, dynamic convolution~\cite{wu2018pay}, and hybrid attentions~\cite{shazeer2019fast,wang2019accelerating}, .etc.

Nonetheless, the above methods still suffer from the inference bottleneck caused by the sequential nature of autoregressive models. Instead, \citet{gu2018nonautoregressive} propose non-autoregressive generation where target words are predicted independently, leading to great speedup, albeit at a high cost to generation quality. Follow-up studies often seek solutions to recover the performance~\cite{libovicky-helcl-2018-end,guo2019non,shao2019minimizing,Ghazvininejad2020AlignedCE,ran-etal-2020-learning}, but also reveal the trade-off between the quality and speed in terms of autoregressiveness. This motivates researchers to discover the optimal balance by resorting to semi-autoregressive modeling~\cite{wang-etal-2018-semi,NIPS2018_8212}, iterative refinement~\cite{lee-etal-2018-deterministic,stern2019insertion,ghazvininejad-etal-2019-mask} or in-between~\cite{pmlr-v80-kaiser18a,akoury-etal-2019-syntactically}.


We hypothesize that generation order affects the felicity of independence assumptions made in semi-autoregressive modelling.
Unlike generation with flexible orders~\cite{NIPS2019_8986,stern2019insertion,tacl_a_00292}, we employ deterministic generation order for model simplicity and training efficiency, specifically focusing on bidirectional decoding.
The study of bidirectional modeling dates back to the era of phase-based statistical machine translation~\cite{watanabe-sumita-2002-bidirectional,finch-sumita-2009-bidirectional} and recently gained popularity in neural machine translation~\cite{liu-etal-2016-agreement-target,sennrich-etal-2016-edinburgh,zhang2019regularizing,8863411,zheng-etal-2019-dynamic}. Unfortunately, these methods either design complex neural decoders, which hurts training efficiency, and/or perform the left-to-right and right-to-left inference separately followed by rescoring, which slows down decoding. By contrast, our model speeds up inference while maintaining training speed.

Our work is closely related to \sa{}~\cite{wang-etal-2018-semi} and synchronous bidirectional generation~\cite{zhou2019sequence}. \bidecoder{} extends \sa{} to incorporate information from different directions. In contrast to \citet{zhou2019sequence}, we only make minimal changes to the standard Transformer decoder, which benefits efficiency during training and inference, and makes our method easy to implement.
We also find improvements in both decoding speed and translation quality compared to~\cite{wang-etal-2018-semi,zhou2019sequence}.

\section{Autoregressive Transformer}

Transformer~\cite{NIPS2017_7181}, the state-of-the-art neural sequence generation model, follows the autoregressive factorization as in Eq. \ref{eq:standard_prob}. To handle the dependency of target word $y_t$ on previous target words $\mathbf{y}_{<t}$, Transformer relies on a masked self-attention network in the decoder:
\begin{equation}\label{eq:masked_san}
    \textsc{Att}(\mathbf{Y}^l, \mathbf{M}) = f\left(\frac{\mathbf{Q}^l{\mathbf{K}^l}^T}{\sqrt{d}} + \mathbf{M}\right)\mathbf{V}^l
\end{equation}
where $\mathbf{Q}^l, \mathbf{K}^l, \mathbf{V}^l = \mathbf{W}_q^l\mathbf{Y}^l, \mathbf{W}_k^l\mathbf{Y}^l,\mathbf{W}_v^l\mathbf{Y}^l \in \mathbb{R}^{n\times d}$, $f(\cdot)$ denotes softmax operation, $d$ is model dimension and $l$ is layer depth. $\mathbf{W}_q, \mathbf{W}_k, \mathbf{W}_v \in \mathbb{R}^{d\times d}$ are trainable parameters.

The mask matrix $\mathbf{M} \in \mathbb{R}^{n\times n}$ limits the access of attention to only the past target words. Formally, given the target sequence length $n$, this matrix can be constructed by the following masking function:
\begin{equation}\label{eq:mask}
    \mathcal{M}_{i,j}(h, c) = 
    \begin{cases}
        0, & \text{if } \ceil*{\nicefrac{i}{(h\cdot c)}} \geq \ceil*{\nicefrac{j}{(h\cdot c)}}\\
        -\infty,      & \text{otherwise}
    \end{cases}.
\end{equation}
where $0<i,j<n$, $h$ denotes the number of generation directions, and $c$ is the number of target words predicted per direction. By default, the Transformer decoder is unidirectional and generates words one-by-one. Thus, $\mathbf{M}=\mathcal{M}(1, 1)$. The infinity here forces softmax output a probability of 0, disabling invalid attentions.

The input layer to Transformer's decoder is the addition of target word embedding $\mathbf{E}_{\mathbf{y}}$ and word position encoding $\text{PE}_{\mathcal{T}}$, i.e $\mathbf{Y}^0 = \mathbf{E}_{\mathbf{y}} + \text{PE}_{\mathcal{T}} \in \mathbb{R}^{n\times d}$. 
$\mathcal{T}$ maps $\mathbf{y}$ to its word position sequence, which is a simple indexing function (Figure \ref{fig:overall:sa}):
\begin{equation}\label{eq:pos_enc}
    \mathcal{T}_t = t-1,
\end{equation}
where $t=1\ldots n$. Transformer adopts the sinusoidal positional encoding to project these indexes to real-space embeddings, and uses the last-layer decoder output $\mathbf{Y}^L$ to predict the respective next target word. 
We explain how to accelerate generation by reordering $\mathbf{y}$, adjusting $h, c$ and $\mathcal{T}$ next.

\section{Interleaved Bidirectional Decoder}

The structure of Transformer is highly parallelizable, but the autoregressive schema $(h=1,c=1)$ blocks this parallelization during inference. We alleviate this barrier by exploring the alternative probabilistic factorization in Eq. \ref{eq:two_head_prob} to allow words predicted from different directions simultaneously. 

We propose \bidecoder{} as shown in Figure \ref{fig:overall:bidecoder}.
We reuse the standard decoder's architecture in a bid to largely inherit Transformer's parallelization and avoid extra computation or parameters, rather than devising dedicated decoder architectures~\cite{zhou2019sequence,ZHANG2020103234}. To make the left-to-right and right-to-left generation collaborative, we reorganize the target sequence and the word positions below (purple and green rectangles in Figure \ref{fig:overall:bidecoder}):
\begin{align}
    \mathbf{y}^{\textsc{bd}} & = \left[y_1 y_n, y_2 y_{n-1}, ..., y_{\floor*{\nicefrac{n}{2}}+1}\right], \label{eq:two_head_seq} \\
    \mathcal{T}^{\textsc{bd}}_t & = (-1)^{(t-1)}\ceil*{\nicefrac{t}{2}}. \label{eq:two_head_pos}
\end{align}
By following the generation order defined by Eq. \ref{eq:two_head_prob}, the sequence $\mathbf{y}^{\textsc{bd}}$ interleaves $\mathbf{y}_{1:\floor*{\nicefrac{n}{2}}}$ and $\mathbf{y}_{\floor*{\nicefrac{n}{2}}+1:n}$ and converts a bidirectional generation problem to a unidirectional one. We introduce negative positions to $\mathcal{T}^{\textsc{bd}}$ to retain the locality bias of sinusoidal positional encodings in $\mathbf{y}^{\textsc{bd}}$.\footnote{Consider Figure \ref{fig:overall:bidecoder}. We cannot reorder position encodings along with embeddings (1,6,2,5,...) because we do not know sentence length at test time. Simply using vanilla position encodings (1,2,3,4,...) would increase the embedding distance between positions within a direction.}
Compared to $(\mathbf{y}, \mathcal{T})$, the reorganized sequences $(\mathbf{y}^{\textsc{bd}}, \mathcal{T}^{\textsc{bd}})$ have the same length, thus with no extra overhead.

We also adapt the self-attention mask to permit step-wise bidirectional generation:
\begin{equation}\label{eq:two_head_mask}
    \mathbf{M}^{\textsc{bd}} = \mathcal{M}(2, 1),
\end{equation}
where \bidecoder{} has $h=2$ generation directions.
This corresponds to the relaxed causal mask by \citet{wang-etal-2018-semi}, which ensures access to all predictions made in previous time steps\footnote{Note that with two tokens produced per time step, decoder inputs are shifted by two.} and allows for interactions among the tokens to be produced per time step.
Although two words are predicted independently at each step, the adapted self-attention mask makes their corresponding decoding context complete; each word has full access to its corresponding decoding history, i.e.\ the left-to-right ($\mathbf{y}_{1:t}$) and right-to-left ($\mathbf{y}_{n-t+1:n}$) context.
Except for $(\mathbf{y}^{\textsc{bd}}, \mathbf{M}^{\textsc{bd}}, \mathcal{T}^{\textsc{bd}})$, other components in Transformer are kept intact, including training objective.

\subsection{Beyond Two-Word Generation}

Eq. \ref{eq:two_head_prob} only supports two-word generation, which indicates an upper bound of 2$\times$ speedup at inference. To improve this bound, we study strategies for multi-word generation. We explore two of them.

\paragraph{Multi-Directional Decoding} Similar to \bidecoder{}, \midecoder{} also permutes the target sequence. It inserts multiple generation directions (i.e.\ increases $h$), with each direction producing one word per step (i.e.\ $c=1$). As shown in Figure \ref{fig:overall:midecoder}, it splits the target sequence into several roughly equal segments followed by applying \bidecoder{} to each segment (thus an even $h$ required).
Formally, \midecoder{} reframes the target sequence and word positions as follows:
\begin{align}
    \mathbf{y}^{\textsc{md}} = \left[\mathbf{y}^{\textsc{bd}}_{1,k}, \mathbf{y}^{\textsc{bd}}_{2,k}, \ldots, \mathbf{y}^{\textsc{bd}}_{\nicefrac{h}{2},k}\right]_{k=1}^{\ceil*{\nicefrac{n}{h}}}, \label{eq:multi_head_seq} \\
    \mathcal{T}^{\textsc{md}}_t = \left(\floor*{\nicefrac{t-1}{h}}, t-1 \text{ mod } h\right), \label{eq:multi_head_pos}
\end{align}
where $\mathbf{y}_{i,k}^{\textsc{bd}}$ denotes the $k$-th word of $\mathbf{y}_i^{\textsc{bd}}$, which is the $i$-th segment of $\mathbf{y}$ reordered by \bidecoder ($\nicefrac{h}{2}$ segments in total). $\mathcal{T}^{\textsc{md}}$ decomposes the word position into two parts. The first one represents the index of decoding step where each word is predicted; the second one denotes the generation direction each target word belongs to. Specifically, we record the corresponding direction indices and add a group of trainable direction embeddings (red rectangles in Figure \ref{fig:overall:midecoder}) into the decoder input.
\midecoder{} uses the following self-attention mask:
\begin{equation}\label{eq:md_mask}
    \mathbf{M}^{\textsc{md}} = \mathcal{M}(h, 1)
\end{equation}

\begin{algorithm}[t]
    \caption{\label{alg:inference} Beam search with step-wise multi-word generation.}
    \begin{algorithmic}[1]
        \Input Decoder \textit{dec}, beam size $B$, word number $z=h\cdot c$, maximum length $T$
        \Output Top-$B$ finished hypothesis
        \Empty{\color{gray} $\triangleright$ {\textit{initial hypothesis ($z$ start symbols, score 0)}}}
        \State $\mathcal{H}_0 \leftarrow \{([\text{`\textit{[s]}'}]^z, 0)\}$        
        \State $\mathcal{H}_{finish} \leftarrow \emptyset$
        \State $t \leftarrow 0$
        \While {$|\mathcal{H}_{finish}| < B \And t < T$}
            \For {$(h_t, s_t) \in \mathcal{H}_t$}
                \Empty \textit{\color{gray}\qquad\quad$\triangleright$ words ${W}_p$ of probability ${P}\in\mathbb{R}^{z\times B}$}
                \State $\mathbf{P}, \mathbf{W}_p \leftarrow top_B(dec(h_t))$ \label{alg:inf:top_dec}
                \Empty \textit{\color{gray} \qquad\quad$\triangleright$ $\oplus$: outer addition for vectors}
                \State $\mathbf{s}, \mathbf{W}_s \leftarrow top_B(\oplus_{i=1}^z \log \mathbf{P}_i)$ \label{alg:inf:top_out_product}
                \Empty \textit{\color{gray} \qquad\quad$\triangleright$ extract words by index, $W\in \mathbb{R}^{B\times z}$}
                \State $\mathbf{W} \leftarrow tracewords(\mathbf{W}_s, \mathbf{W}_p)$ \label{alg:inf:tracewords}
                
                \For {$(\mathbf{w}, s) \textbf{ in } (\mathbf{W}, \mathbf{s})$}
                    \Empty \textit{\color{gray}\qquad\qquad~ $\triangleright$ meet end-of-hypothesis condition}
                    \If {\textit{finish}($\mathbf{w}$)} \label{alg:inf:eos}
                        \State add $([h_t, \mathbf{w}], s+s_t)$ to $\mathcal{H}_{finish}$
                    \Else
                        \State add $([h_t, \mathbf{w}], s+s_t)$ to $\mathcal{H}_{t+z}$
                    \EndIf               
                \EndFor
            \EndFor
            \State prune $\mathcal{H}_{t+z}$ to keep top-$B$ hypothesis
            \State $t \leftarrow t + z$
        \EndWhile
        \Empty \textit{\color{gray} $\triangleright$ post($\cdot$): process $h_t$ to recover word order}
        \State\Return sort $(post(h_t), s_t) \in \mathcal{H}_{finish}$ by $\frac{s_t}{t}$ \label{alg:inf:return}
    \end{algorithmic}
\end{algorithm}

\paragraph{Semi-Autoregressive Decoding} Instead of partitioning the target sequence,
another option is to produce multiple target words per direction at each step~\cite[i.e.\ increase $c$,][]{wang-etal-2018-semi}. \sa{} assumes that neighbouring words are conditionally independent, despite the fact that tokens in natural language are typically highly inter-dependent.

We combine \sa{} with \bidecoder{} (Figure \ref{fig:overall:bidecoder_sa}) with the expectation that producing 2 neighbouring tokens independently per direction is less harmful than producing 4 neighbouring words in parallel.
We reuse the sequence $\mathbf{y}^{\textsc{bd}}$ and $\mathcal{T}^{\textsc{bd}}(n)$ for the decoder input, but enlarge the attention range in the self-attention mask to assist multi-word generation (Figure \ref{fig:overall:mask4}):
\begin{equation}\label{eq:sa_mask}
\mathbf{M}^{\textsc{sa}} = \mathcal{M}(2, c).
\end{equation}

\subsection{Inference}

To handle multiple predicted words per decoding step simultaneously, we adjust the beam search algorithm as in Algorithm \ref{alg:inference}. For each partial hypothesis $h_t$, we predict $z=h\cdot c$ words in parallel. We thus first extract the $B$ top-scoring predictions $\mathbf{W}_p$ of probability $\mathbf{P}$ for all $z$ positions (line \ref{alg:inf:top_dec}), followed by pruning the resulting search space of size $\mathcal{O}(B^z)$ through an outer-addition operation to size $B$ (line \ref{alg:inf:top_out_product}). 
The scores $\mathbf{s} \in \mathbb{R}^{B}$ (line \ref{alg:inf:top_out_product}) and the backtraced words $\mathbf{W} \in \mathbb{R}^{B \times z}$ (line \ref{alg:inf:tracewords}) are then used for normal decoding. Note that each complete hypothesis requires a simple deterministic post-processing to recover its original word order (line \ref{alg:inf:return}). In contrast to \citet{zhou2019sequence}, we do not separate the left-to-right beam from the right-to-left beam.

\paragraph{End-of-Hypothesis Condition} With multiple predicted target words, determining whether one hypothesis is complete or not becomes challenging. We adopt a simple strategy: one hypothesis is assumed complete once any word in the predictions hits the end-of-sentence symbol (``[/s]'') (line \ref{alg:inf:eos}).
We leave the study of alternatives for the future.

\begin{table*}[t]
    \centering
    \small
    \begin{tabular}{llcccccccc}
      \toprule
      ID & Model & $B$ & $h$ & $c$ & BLEU$\uparrow$ & +KD$\uparrow$ & Latency$\downarrow$ & Speedup$\uparrow$ & Train$\uparrow$ \\
      \midrule
      \multirow{2}{*}{1} & \multirow{2}{*}{Transformer} & 4 & \multirow{2}{*}{1} & \multirow{2}{*}{1} & 26.9 & 27.3 & 387 & 1.00$\times$ & \multirow{2}{*}{1.00$\times$} \\
      & & 1 & & & 26.0 & 26.8 & 294 & 1.32$\times$ &  \\
      \midrule
      \multirow{2}{*}{2} & \multirow{2}{*}{\bidecoder{}} & 4 & \multirow{2}{*}{2} & \multirow{2}{*}{1} & 26.2 & 27.1 & 204 & 1.90$\times$ & \multirow{2}{*}{0.98$\times$}\\
      & & 1 & & & 25.0 & 26.8 & 166 & 2.33$\times$ \\
      \midrule
      \multirow{2}{*}{3} & \multirow{2}{*}{2 + \sa} & 4 & \multirow{2}{*}{2} & \multirow{2}{*}{2} & 23.0 & 26.3 & 117 & 3.31$\times$ & \multirow{2}{*}{0.98$\times$}  \\
      & & 1 & & & 21.7 & 26.0 & 89 & 4.35$\times$ \\
      \midrule
      \multirow{2}{*}{4} & \multirow{2}{*}{\midecoder{}} & 4 & \multirow{2}{*}{4} & \multirow{2}{*}{1} & 21.5 & 24.6 & 102 & 3.79$\times$ & \multirow{2}{*}{0.98$\times$} \\
      & & 1 & & & 19.7 &  24.1 & 85 & 4.55$\times$  \\
      \bottomrule
    \end{tabular}
    \caption{Performance on WMT14 En-De for different models with respect to beam size ($B$), generation direction number ($h$, Eq. \ref{eq:mask}) and predicted token number per step ($c$, Eq. \ref{eq:mask}). \textit{BLEU}: detokenized BLEU for models trained from scratch, \textit{+KD}: detokenized BLEU for models trained with knowledge distillation. \textit{Latency} (in millisecond) and \textit{Speedup} are evaluated by decoding the test set with a batch size of 1, averaged over three runs. We report the latency and speedup for \textcircled{2}, \textcircled{3} and \textcircled{4} trained with KD. \textit{Train} compares the training speed averaged over 100 steps. Time is measured on GeForce GTX 1080.} 
    \label{tab:wmt14_ende}
\end{table*}
\section{Experiments}

\paragraph{Setup}
We test our model on machine translation (MT) and document summarization. We train MT models on five different language pairs: WMT14 English-German~\cite[En-De,][]{bojar-EtAl:2014:W14-33}, WMT14 English-French~\cite[En-Fr,][]{bojar-EtAl:2014:W14-33}, WMT16 Romanian-English~\cite[Ro-En,][]{bojar-etal-2016-findings}, WMT18 English-Russian~\cite[En-Ru,][]{bojar-etal-2018-findings} and WAT17 Small-NMT English-Japanese~\cite[En-Ja,][]{nakazawa-etal-2017-overview}. Translation quality is measured by BLEU~\cite{papineni-etal-2002-bleu}, and we report detokenized BLEU using the toolkit \textit{sacreBLEU}~\cite{post-2018-call}\footnote{Signature BLEU+c.mixed+\#.1+s.exp+tok.13a+v.1.4.3} except for En-Ja. Following~\citet{NIPS2019_9297}, we segment Japanese text with KyTea\footnote{http://www.phontron.com/kytea/} and compute tokenized BLEU. We train document summarization models on two benchmark datasets: the non-anonymized version of the CNN/Daily Mail dataset~\cite[CDMail,][]{NIPS2015_5945} and the Annotated English Gigaword~\cite[Gigaword,][]{rush-etal-2015-neural}. We evaluate the summarization quality using ROUGE-L~\cite{lin-2004-rouge}. 


We provide details of data preprocessing and model settings in Appendix \ref{sec:app:model_setting}.
We perform thorough analysis of our model on WMT14 En-De. We also report results improved by knowledge distillation~\cite[KD,][]{kim-rush-2016-sequence}.

\subsection{Results on WMT14 En-De}

Table \ref{tab:wmt14_ende} compares the performance of our models on WMT14 En-De. Relaxing the autoregressiveness with \bidecoder{} yields slightly worse translation quality compared to Transformer (-0.7 BLEU, \textcircled{1}$\rightarrow$\textcircled{2}, w/o KD, $B=4$). Unlike \citet{ZHANG2020103234}, we observe no quality improvement, but our model delivers a speedup of 1.90$\times$$\sim$2.33$\times$ at inference, clearly surpassing the simple greedy decoding baseline (1.32$\times$) and BIFT (0.89$\times$)~\cite{ZHANG2020103234}. The dropped quality is easily recovered with knowledge distillation (+0.2 BLEU, \textcircled{1}$\rightarrow$\textcircled{2}, w/ KD, $B=4$).

Going beyond two-word generation, which enhances independence, greatly decreases the performance (\textcircled{2}$\rightarrow$\textcircled{3},\textcircled{4}, w/o KD) while enlarging the speedup to 3.3$\times$--4.5$\times$. Compared to \sa{}, the quality degradation with \midecoder{} is larger, both w/ and w/o KD. We ascribe this to the difficulty of structure planning, as \midecoder{} has to guess words in the middle of the sequence at the start of generation. We employ \sa{} for the following experiments.

In contrast to existing work~\cite{zhang2018asynchronous,8863411,ZHANG2020103234,zhou2019sequence}, our models marginally affect the training efficiency (0.98$\times$ vs 0.61$\times$~\cite{ZHANG2020103234}), and require no extra linguistic information~\cite{akoury-etal-2019-syntactically}. Our results also suggest that the degree each model benefits from KD varies. Follow-up studies should report performance w/ and w/o KD.

\begin{table}[t]
    \centering
    \small
    \begin{tabular}{llccc}
      \toprule
      ID & Model & $h$ & $c$ & BLEU$\uparrow$ \\
      \midrule
      1 & \bidecoder{} & 2 & 1 & 26.2 \\
      \midrule
      2 & 1 + vanilla mask & 2 & 1 & 25.7 \\
      3 & 1 + vanilla positions & 2 & 1 & 25.9 \\
      4 & 1 + middle-to-side & 2 & 1 & 20.7 \\
      5 & 1 + indep. directions & 2 & 1 &  23.9 \\
      6 & vanilla \sa{} & 1 & 2 & 24.1 \\
      \midrule
      7 & 1 + \sa{} & 2 & 2 & 23.0 \\
      8 & vanilla \sa{} & 1 & 4 & 18.7 \\
      \bottomrule
    \end{tabular}
    \caption{Ablation study on WMT14 En-De. Beam size 4. All models are trained from scratch. \textit{vanilla mask}/\textit{vanilla positions}: the self-attention mask ($\mathcal{M}(1,1)$, Eq. \ref{eq:mask}) and word positions ($\mathcal{T}$, Eq. \ref{eq:pos_enc}) used in Transformer. \textit{middle-to-side}: generate words from the middle of the sequence to its two ends, a reverse mode of \bidecoder{}. \textit{indep. directions}: disable cross-direction interaction. \textit{vanilla \sa{}}: predict multiple target words per step following one direction~\cite{wang-etal-2018-semi}. 
    }
    \label{tab:wmt14_ende_ablation}
\end{table}

\begin{table}[t]
    \centering
    \small
    \begin{tabular}{ccc}
      \toprule
      & Left-to-Right & Bidirectional \\
      \midrule
      Autoregressive & 4.04 & 4.86  \\
      Semi-Autoregressive & 6.95 & 4.72 \\
      \midrule
      Estimated PMI & 0.235 & -0.014 \\
      \bottomrule
    \end{tabular}
    \caption{Perplexity of autoregressive and semi-autoregressive models with different factorizations, and estimated average point-wise mutual information between words that are predicted independently. Measured on WMT14 En-De test set. \textit{Left-to-Right}: $h=1$, \textit{Bidirectional}: $h=2$; Autoregressive: $z=1$, Semi-autoregressive: $z=2$. The estimated PMI shows that the inter-dependence of word pairs predicted in parallel by vanilla \sa{} is stronger than for those predicted simultaneously by \bidecoder{}.}
    \label{tab:wmt14_ende_mi}
\end{table}

\paragraph{Ablation Study} We carry out an ablation study as shown in Table \ref{tab:wmt14_ende_ablation}. Replacing the attention mask with the vanilla one (\textcircled{1}$\rightarrow$\textcircled{2}) introduces unnecessary independence assumptions and  reduces performance by 0.5 BLEU. Using vanilla positional encodings (\textcircled{3}) also reduces performance -0.3 BLEU, indicating that we benefit from preserving the locality bias of sinusoidal encodings within each direction.
 Changing the generation direction from the side-to-middle (\textcircled{1}) to the middle-to-side (\textcircled{4}) dramatically increases the learning difficulty (-5.5 BLEU).
 
  In \bidecoder{}, the two translation directions are interlinked, i.e.\ predictions are conditioned on the history of both directions. We can remove cross-direction attention, essentially forcing the model to produce the left and right half of sequences independently. Such an independent generation performs poorly (-2.3 BLEU, \textcircled{1}$\rightarrow$\textcircled{5}), which supports the importance of using bidirectional context and resonates with the finding of \citet{zhou2019sequence}. 

\paragraph{Vanilla \sa{} vs. \bidecoder{}} 
Our \bidecoder{} shares architectural properties with vanilla \sa{}~\cite{wang-etal-2018-semi}, namely the independent generation of two tokens per time step, and the adapted self-attention mask, but crucially differ in their generation order and independence assumptions, with vanilla \sa{} operating from left-to-right, and \bidecoder{} interleaving left-to-right and right-to-left decoding.

Our ablation results in Table \ref{tab:wmt14_ende_ablation} show that \bidecoder{} substantially outperforms vanilla \sa{} (2.1/4.3 BLEU, \textcircled{1}$\rightarrow$\textcircled{6}/\textcircled{7}$\rightarrow$\textcircled{8}).
To further investigate the difference in independence assumptions between the two approaches, we compare estimated point-wise mutual information (PMI) of the words being predicted independently by \bidecoder{} and vanilla \sa{}.\footnote{Details about PMI estimation are given in Appendix \ref{sec:app:pmi}} Results in Table \ref{tab:wmt14_ende_mi} show that the PMI in \bidecoder{} ($-0.014$) is significantly smaller than that in vanilla \sa{} ($0.235$), supporting our assumption that distant words are less inter-dependent on average. This also explains the smaller quality loss in \bidecoder{} compared to vanilla \sa{}.

\begin{table}[t]
    \centering
    \small
    \begin{tabular}{lccc}
      \toprule
      Model & $L/h/c$ & BLEU$\uparrow$ & Speedup$\uparrow$ \\
      \midrule
      Transformer & 6/1/1 & 26.9 & 1.00$\times$ \\
      \quad + student & 2/1/1 & 26.0 & 2.19$\times$ \\
      \quad + KD & 2/1/1 & 26.7 & 2.32$\times$ \\
      \midrule
      \bidecoder{} & 6/2/1 & 26.2 & 1.90$\times$ \\
      \quad + student & 2/2/1 & 25.0 & 4.29$\times$ \\
      \quad + KD & 2/2/1 & 26.6 & 4.41$\times$ \\
      \midrule
      \bidecoder{} + SA & 6/2/2 & 23.0 & 3.31$\times$ \\
      \quad + student & 2/2/2 & 21.5 & 7.13$\times$ \\
      \quad + KD & 2/2/2 & 24.5 & 7.24$\times$ \\
      \bottomrule
    \end{tabular}
    \caption{Detokenized BLEU and decoding speedup for student models on WMT14 En-De with reduced decoder depth $L$ (encoder depth remains constant). Beam size 4.}
    \label{tab:wmt14_ende_teacher_student}
\end{table}

\paragraph{On Teacher-Student Model} One classical approach to improving decoding efficiency is  training a small student model w/ KD. Results in Table \ref{tab:wmt14_ende_teacher_student} support this: Transformer with a student model produces similar performance w/ KD but runs 2.32$\times$ faster, even better than \bidecoder{} (1.90 $\times$). Combining the student schema with \bidecoder{} increases the speedup to 4.41$\times$ without hurting the performance (26.6 BLEU, w/ KD). In exchange of 2.4 BLEU, we could reach 7.24$\times$ faster decoding with \sa{}. The compatibility of our model with the teacher-student framework reflects the generalization of our bidirectional modeling.
The results also demonstrate that efficiency improvements from faster autoregressive decoding, here obtained by reducing the number of decoder layers $L$\footnote{Also note the concurrent work by \citep{kasai2020deep}.}, and from bidirectional decoding, are orthogonal.

\begin{figure}[t]
  \centering
    \includegraphics[scale=0.45]{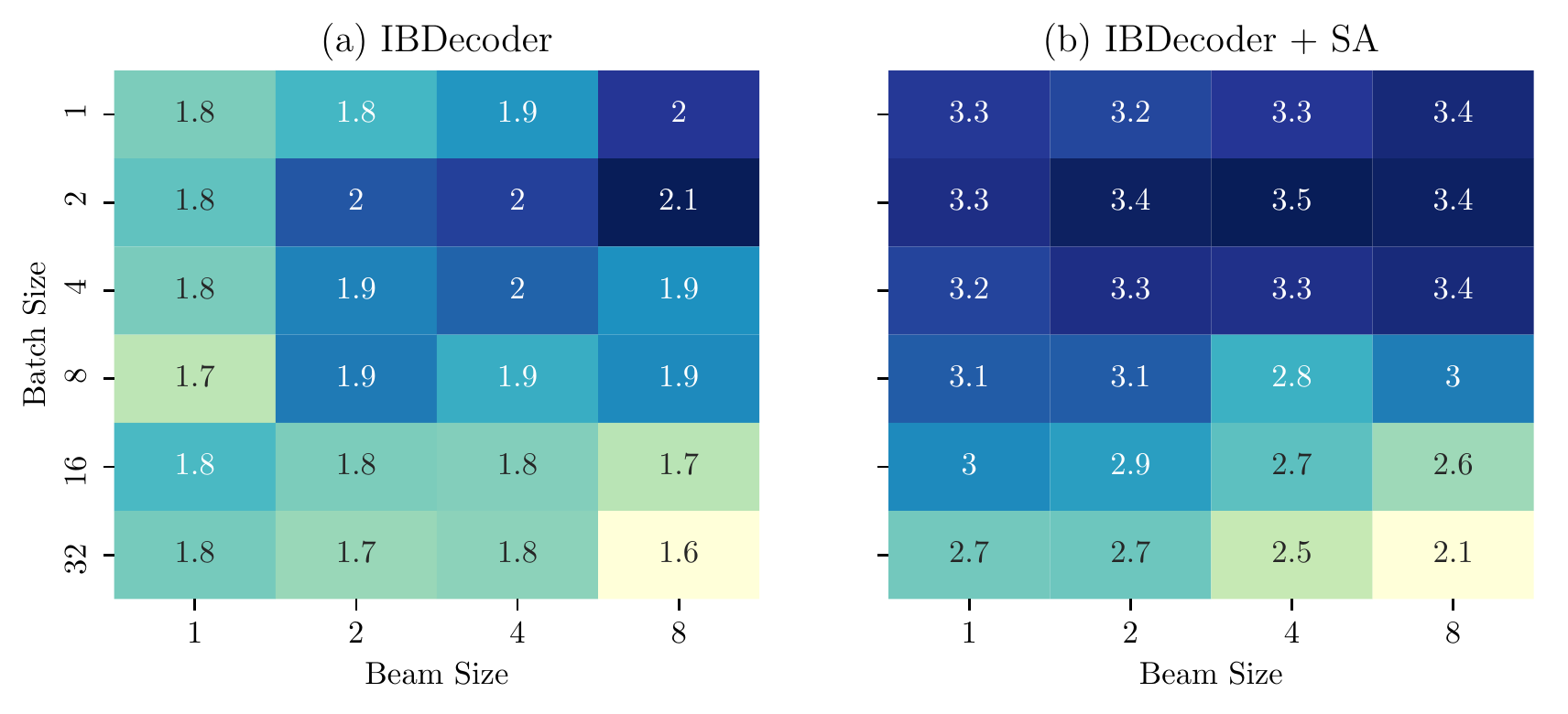}
  \caption{\label{fig:batch_beam} Speedup against Transformer vs. batch size and beam size on WMT14 En-De. Comparison is conducted under the same batch size and beam size. \bidecoder{} (+\sa{}) is trained with KD. Our model consistently accelerates decoding.}
\end{figure}

\paragraph{Impact of Batch and Beam Size} Figure \ref{fig:batch_beam} shows speedups over a standard Transformer with varying batch and beam sizes. When batch size $<8$, increasing beam size improves the speedup; while the impact becomes negative with batch size $\geq 8$. Overall, our model is consistently faster than Transformer at inference, regardless of the batch and beam size.

\begin{figure}[t]
  \centering
    \includegraphics[scale=0.42]{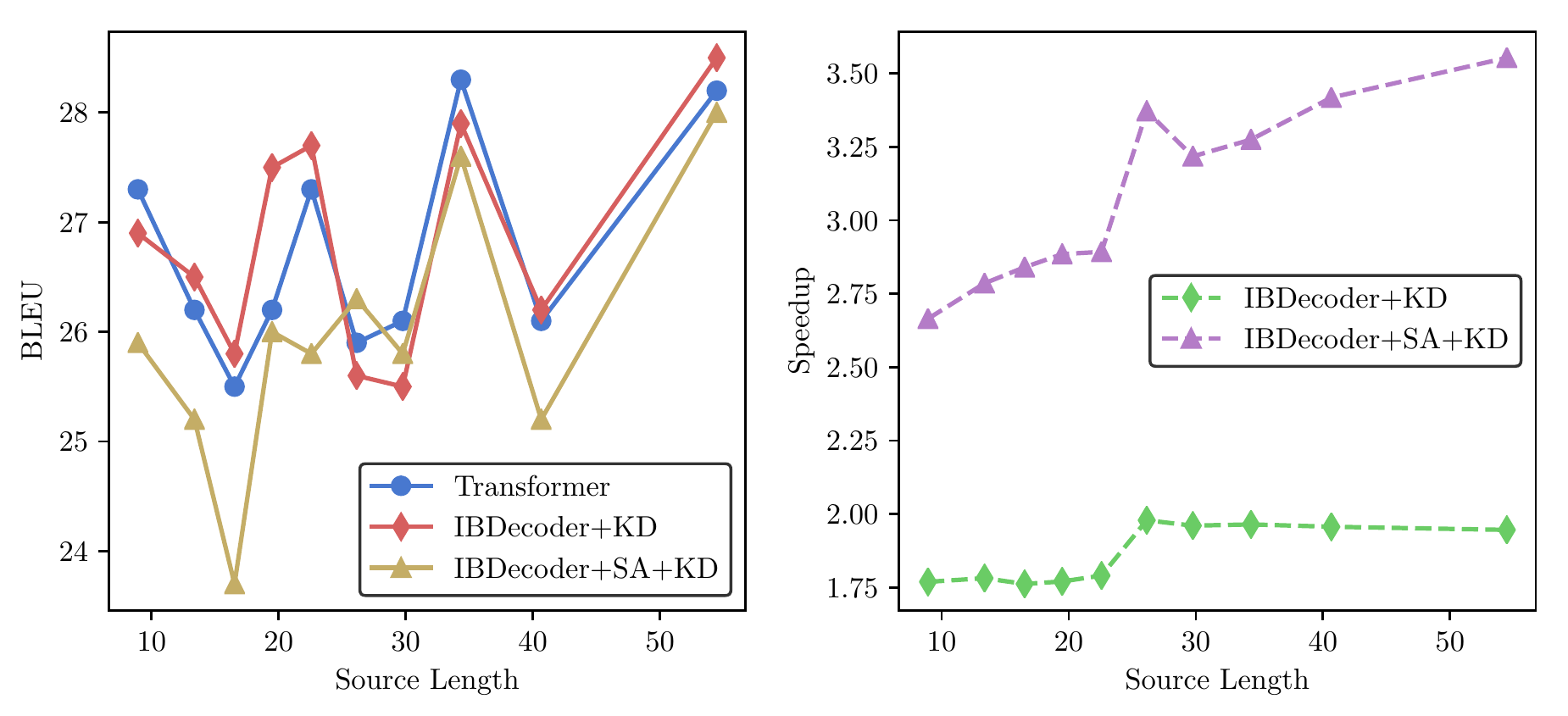}
  \caption{\label{fig:length} BLEU (solid lines, left) and speedup (dashed lines, right) as a function of source sentence length on WMT14 En-De. We sort the test set according to the source sentence length and uniformly divide it into 10 bins (274 sentences each). \bidecoder{} (+\sa{}) is trained with KD. Beam size 4.}
\end{figure}

\paragraph{Impact of Source Sentence Length} Although translation quality fluctuates over the source sentence length, Figure \ref{fig:length} shows that our model shares the same performance pattern with the baseline. With respect to the speedup, our model performs better when translating longer source sentences.

\begin{figure}[t]
  \centering
    \includegraphics[scale=0.42]{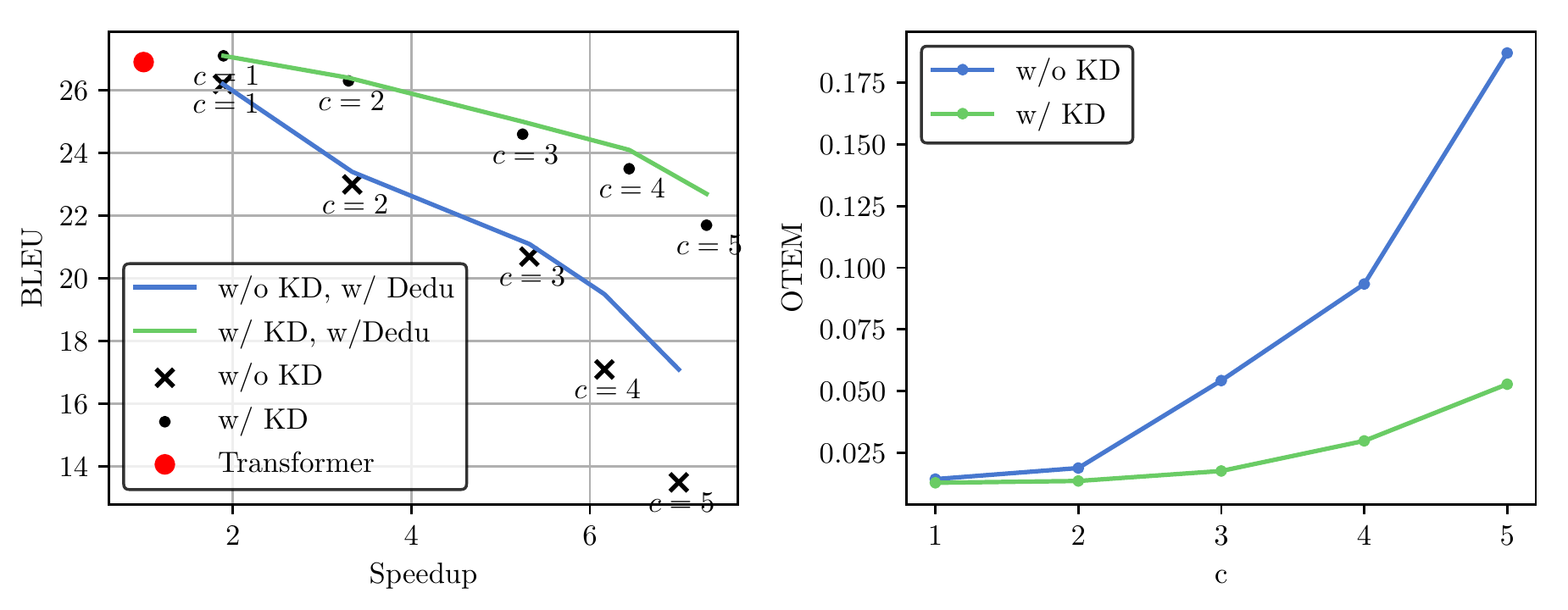}
  \caption{\label{fig:impact_c} BLEU versus speedup (left) and OTEM (right) for different $c$ on WMT14 En-De. Generation directions: $h=2$. Beam size 4. OTEM$\downarrow$: a metric measuring the degree of over-translation~\cite{yang2018otem}. Larger $c$ indicates more independence between neighbouring tokens and results in more severe over-translation. Deduplication (Dedu) improves translation quality for large $c$.}
\end{figure}

\paragraph{Effect of $c$} Results in Figure \ref{fig:impact_c} show that $c$ controls the trade-off between translation quality and speedup. With larger $c$, more target tokens are predicted per decoding direction, leading to better speedup, but causing a larger performance drop w/ and w/o KD. Further analysis reveals that, as the dependency between predicted target words weakens, our model suffers from more serious over-translation issue, yielding larger OTEM~\cite{yang2018otem}. Although n-gram deduplication slightly improves quality\footnote{we only applied deduplication for results in Figure \ref{fig:impact_c}.}, it does not explain the whole performance drop, echoing with~\citet{wang-etal-2018-semi}. We recommend using $c=2$ for a good balance. In addition, the reduction of OTEM by KD in Figure \ref{fig:impact_c} partially clarifies its improvement on quality.

\begin{figure}[t]
  \centering
    \includegraphics[scale=0.50]{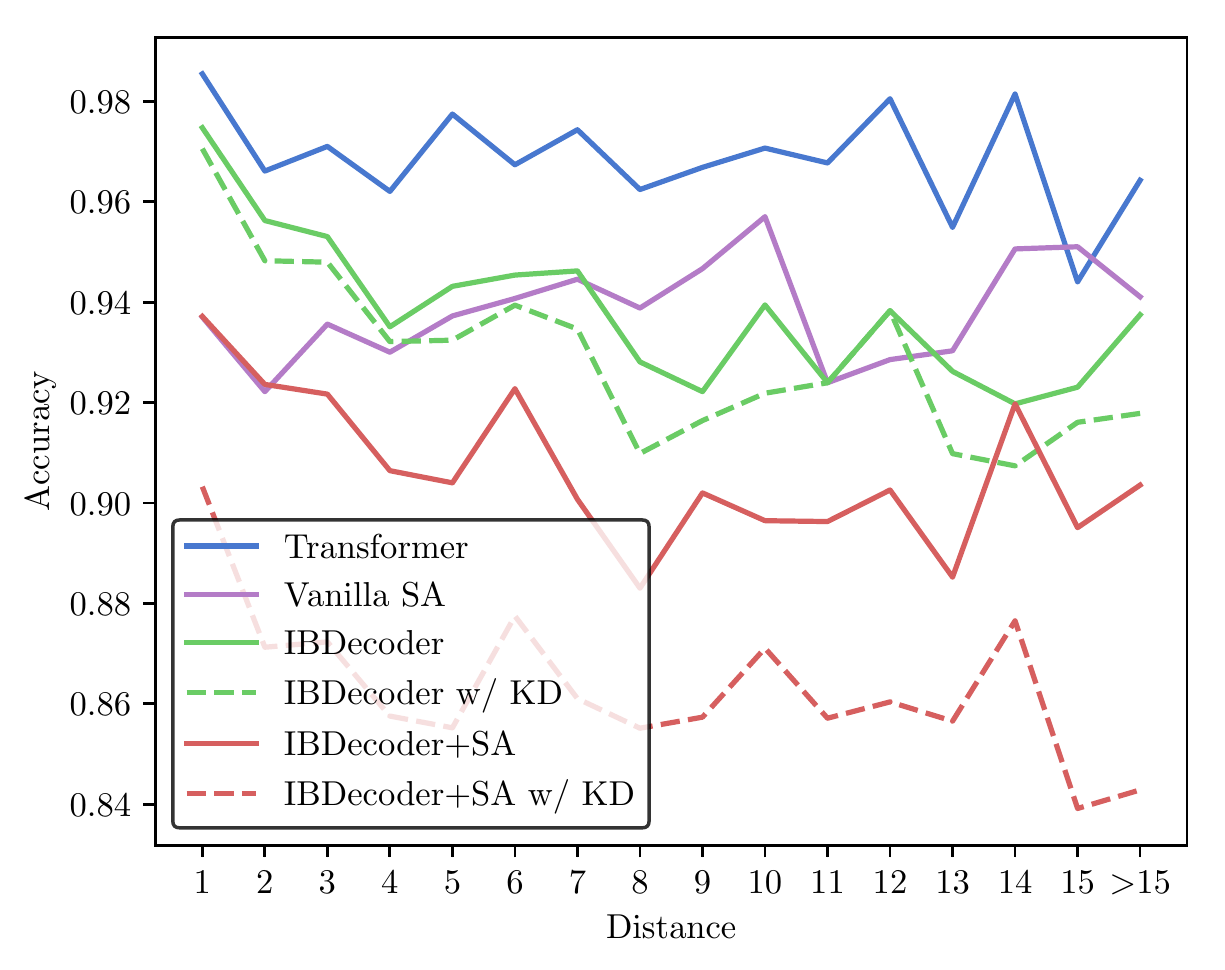}
  \caption{\label{fig:subject_verb} Accuracy of different models over distances on the subject-verb agreement task in \textit{Lingeval97}.}
\end{figure}

\begin{table}[t]
    \centering
    \small
    \resizebox{\columnwidth}{!}{
    \begin{tabular}{lrcc}
      \toprule
      \multicolumn{2}{l}{Model} & BLEU$\uparrow$ & SU$\uparrow$ \\
      \midrule
      \multicolumn{2}{l}{Existing work} \\
      \multicolumn{2}{l}{SAT~\cite{wang-etal-2018-semi}$^*$} & 26.09$^\dagger$ & 2.07 $\times$ \\
      \multicolumn{2}{l}{SBSG~\cite{zhou2019sequence}$^*$} & 27.22$^\dagger$ & 1.61 $\times$ \\
      \multicolumn{2}{l}{SynST~\cite{akoury-etal-2019-syntactically}} & 20.74 & 4.86$\times$ \\
      \multicolumn{2}{l}{Levenshtein~\cite{NIPS2019_9297}$^*$} & 27.27$^\dagger$ & 4.01$\times$ \\
      \multicolumn{2}{l}{CMLM~\cite{ghazvininejad-etal-2019-mask}$^*$} & 27.03$^\dagger$ & - \\
      \multicolumn{2}{l}{AXE~\cite{Ghazvininejad2020AlignedCE}$^*$} & 23.53$^\dagger$ & - \\
      \midrule
      This work & SacreBLEU$\uparrow$ \\
      {\bidecoder{}}  & 25.0 & 25.73$^\dagger$ & {2.48$\times$} \\
      {\quad w/ \sa{}}  & 22.3$^\diamond$ & 22.95$^\dagger$ & {4.53$\times$} \\
      \quad w/ student & 25.0 & 25.33$^\dagger$ & 4.29$\times$ \\
      {\bidecoder{}$^*$} & 26.8 & 27.50$^{\dagger}$ & {2.33$\times$} \\
      {\quad w/ \sa{}$^*$} & 26.0$^\diamond$ & 26.84$^{\dagger\diamond}$ & {4.35$\times$} \\
      \quad w/ student$^*$ & 26.6 & 27.00$^{\dagger}$ & 4.41$\times$ \\
    \bottomrule
    \end{tabular}}
    \caption{Comparison to several recent fast sequence generation models on WMT14 En-De. $^*$: trained w/ KD. $^\dagger$: tokenized BLEU. $^\diamond$: deduplication applied. \textit{SU}: short for speedup.}
    \label{tab:wmt14_ende_comparison}
\end{table}

\paragraph{Analysis on Long-range Dependency} We adopt the subject-verb agreement task from \textit{Lingeval97}~\cite{sennrich-2017-grammatical} for analysis.
We can see from the results in Figure \ref{fig:subject_verb} that \bidecoder{} performs similarly to the original Transformer for agreement over short distances, but agreement over longer distances drops on average.
In contrast, models that include \sa{} show steep drops in accuracy for short distances.

Curiously, KD seems to harm agreement scores even though it led to higher BLEU. 
Overall, these results suggest that BLEU does not show the full quality loss incurred by our independence assumptions. This deficiency also provides evidence for the performance drop in Figure \ref{fig:impact_c}.

\paragraph{Comparison to Previous Work}

Results in Table \ref{tab:wmt14_ende_comparison} show that our model outperforms SynST~\cite{akoury-etal-2019-syntactically} in quality, and slightly surpasses the Levenshtein Transformer~\cite{NIPS2019_9297} in speed. Particularly, our model ($27.50^\dagger/2.33\times$) surpasses SAT~\cite{wang-etal-2018-semi} ($26.09^\dagger/2.07\times$) and SBSG~\cite{zhou2019sequence} ($27.22^\dagger/1.61\times$) in terms of both quality and speed. Our model doesn't heavily rely on extra linguistic knowledge~\cite{akoury-etal-2019-syntactically}, neither requires complex pseudo training data construction~\cite{NIPS2019_9297}. Compared to these prior studies, our approach is simple but effective.

\begin{table*}[t]
    \centering
    \small
    \resizebox{\textwidth}{!}{
    \begin{tabular}{cllccccccc}
      \toprule
      \multirow{2}{*}{$B$}& & \multirow{2}{*}{Model} & \multirow{2}{*}{KD} & \multicolumn{4}{c}{Machine Translation} & \multicolumn{2}{c}{Document Summarization}\\\cmidrule(lr){5-8} \cmidrule(lr){9-10}
       & &  & & En-Fr & Ro-En & En-Ru & En-Ja & Gigaword & CDMail \\
      \midrule
      \multirow{7}{*}{4} & \multirow{5}{*}{Quality$\uparrow$} &
        Transformer & no & 32.1 & 32.7 & 27.7 & \textbf{43.97} & 35.03 & 36.88 \\ 
      & & \bidecoder{} & no & 32.1 & 33.3 & 27.0 & 43.51 & 34.57 & 36.11 \\ 
      & & \quad + \sa{} & no  & 30.3 & 31.3 & 25.0 & 41.75 & 33.65 & 35.27 \\
      & &  \bidecoder{} & yes & \textbf{32.7} & \textbf{33.5} & 27.5 & 43.76 & 35.12 & 36.46 \\
      & & \quad + \sa{} & yes & 31.3 & 32.7 & 26.4 & 42.99 & 34.74 & 36.27 \\
      \cmidrule{2-10}
      & Latency$\downarrow$ & 
      \bidecoder{} & yes & 231/1.75$\times$ & 205/1.79$\times$ & 204/1.82$\times$ & 157/1.86$\times$ & 83/2.35$\times$ & 657/3.02$\times$\\
      & {/Speedup$\uparrow$} & \quad +\sa{} & yes  & 119/3.41$\times$  & 109/3.37$\times$ & 112/3.30$\times$ & 94/3.10$\times$ & 47/4.20$\times$ & 303/6.55$\times$ \\
      \midrule
      \multirow{8}{*}{1} & \multirow{5}{*}{Quality$\uparrow$} &
        Transformer & no & 31.6 & 32.3 & 27.8 & 42.95 & 34.88 & 34.51 \\ 
      & & \bidecoder{} & no & 31.7 & 32.6 & 26.8 & 43.29 & 34.22 & 36.74 \\ 
      & & \quad + \sa{}    & no   & 29.0 & 30.4 & 24.3 & 41.05 & 33.25 & 35.04 \\
      & &  \bidecoder{} & yes & 32.2 & 33.2 & \textbf{28.2} & 43.79 & \textbf{35.18} & \textbf{37.03} \\
      & & \quad + \sa{} & yes & 30.7 & 32.4 & 26.5 & 42.70 & 34.63 & 36.39 \\
      \cmidrule{2-10}
      & Latency$\downarrow$ & 
          Transformer & no & 357/1.14$\times$ & 333/1.10$\times$ & 342/1.09$\times$ & 260/1.12$\times$ & 157/1.24$\times$ & 1447/1.37$\times$\\
      &\multirow{2}{*}{/Speedup$\uparrow$} & \bidecoder{} & yes & 186/2.18$\times$ & 154/2.37$\times$ & 157/2.37$\times$ & 121/2.40$\times$ & 56/3.51$\times$ & 312/6.36$\times$\\
      & & \quad +\sa{}  & yes & 96/4.20$\times$ & 88/4.17$\times$ & 90/4.14$\times$ & 67/4.34$\times$ & 34/5.83$\times$ & 178/11.15$\times$\\
      \bottomrule
    \end{tabular}}
    \caption{Generation quality (BLEU for MT, Rouge-L for summarization) and latency(ms)/speedup on different tasks. We compare \bidecoder{} (+\sa{}) with Transformer. Best quality is in \textbf{bold}.}
    \label{tab:other_tasks}
\end{table*}

\subsection{Results on Other Tasks}

Table \ref{tab:other_tasks} shows MT results for other translation directions, and for document summarization. Regardless of syntactic, morphological, transcript and sequence-length differences, our model achieves comparable generation quality and 1.75$\times$--11.15$\times$ speedup over different tasks. With KD, our model even outperforms the Transformer baseline on 5 out of 6 tasks. In particular, our model succeeds on the CDMail task which previous non-autoregressive models rarely attempt due to its lengthy target sequence, although our model suffers from the long-range dependency issue as in Figure \ref{fig:subject_verb}.

\section{Conclusion and Future Work}

We present interleaved bidirectional sequence generation to accelerate decoding by enabling generation from the left-to-right and right-to-left directions simultaneously. We combine the strengths of SBSG~\cite{zhou2019sequence} and \sa{}~\cite{wang-etal-2018-semi}, and propose a simple interleaved bidirectional decoder (\bidecoder{}) that can be easily implemented on top of a standard unidirectional decoder, like Transformer, via interleaving the target sequence and tweaking the word positions and self-attention masks. \bidecoder{} inherits Transformer's training parallelization with no additional model parameters, and is extensible with \sa{} and multi-directional decoding.
We show that the independence assumptions we introduce between the two directions are less harmful to translation quality than the independence assumptions in left-to-right \sa{}.
On a series of generation tasks, we report comparable quality with significant inference speedup (4$\times$--11$\times$) and little training overhead.
We also show that the approach is orthogonal to speedups to autoregressive decoding, e.g. by reducing model size.

In the future, we would like to further improve multi-directional generation, and will investigate alternative ways to partition the target sequence and encode positional information. 
We are also interested in better measuring and reducing the quality loss resulting from long-distance dependencies.
Finally, we would like to adapt our interleaving approach to other sequence-to-sequence architectures.

\section*{Acknowledgments}

This work was performed using resources provided by the Cambridge Service for Data Driven Discovery (CSD3) operated by the University of Cambridge Research Computing Service (\url{http://www.csd3.cam.ac.uk/}), provided by Dell EMC and Intel using Tier-2 funding from the Engineering and Physical Sciences Research Council (capital grant EP/P020259/1), and DiRAC funding from the Science and Technology Facilities Council (\url{www.dirac.ac.uk}).
Ivan Titov acknowledges
support of the European Research Council (ERC Starting grant 678254) and the Dutch National Science Foundation
(NWO VIDI 639.022.518).
Rico Sennrich acknowledges support of the Swiss National Science Foundation (MUTAMUR; no.\ 176727).


\bibliographystyle{acl_natbib}
\bibliography{emnlp2020}


\appendix

\section{Data Preprocessing and Model Settings}\label{sec:app:model_setting}

We use the given well-processed data for WAT17 En-Ja. For other tasks, we apply the byte pair encoding model~\cite{sennrich-etal-2016-neural} with a joint vocab size of 32K except for WMT18 En-Ru (48K). We experiment with Transformer Base~\cite{NIPS2017_7181}: $d=512$, $L=6$, 8 attention heads and FFN size of 2048. Dropout of rate 0.1 is used on residual connections and attention weights. We employ Adam $(\beta_1=0.9, \beta_2=0.98)$~\cite{kingma2014adam} for parameter optimization with a scheduled learning rate of warm-up step 4K. Gradient is estimated over roughly 25K target subwords. We average the last 5 checkpoints for evaluation, and use beam search (beam size 4, length penalty 0.6) by default for inference. 

\section{Estimation of the PMI}\label{sec:app:pmi}

To evaluate the average point-wise mutual information (PMI) in Table \ref{tab:wmt14_ende_mi}, we compare \bidecoder{}/vanilla \sa{} to its autoregressive counterpart in terms of testing perplexity (ppl). Take \sa{} ($h=1, c=2$) as example, we have:
\begin{align}
    \text{PMI}(\sa{}) = \log \text{ppl}(\sa{}) - \log \text{ppl}(\text{Base})
\end{align}
where \textit{Base} denotes the baseline Transformer. The intuition behind our estimation is that Transformer handles neighboring words ($y_1, y_2$) autoregressively, thus models their joint probability: $p(y_1, y_2) = p(y_1) \cdot p(y_2|y_1)$. Instead, vanilla \sa{} predicts those words independently, i.e. $p(y_1) \cdot p(y_2)$. Comparing the perplexity of \sa{} and Transformer gives an estimation of the average PMI. The method for \bidecoder{} follows the same spirit.


\end{document}